\documentclass{mva_style}
\usepackage{graphicx}
\usepackage{mathrsfs}
\usepackage{parskip}
\usepackage{indentfirst}

\usepackage{cite}

\usepackage{amsmath}
\usepackage{amssymb}

\finalcopy 

\begin{document}

\title{DCNN-GAN: Reconstructing Realistic Image from fMRI}

\author{
\and
  Yunfeng Lin, Jiangbei Li, Hanjing Wang\\
  Shanghai Jiao Tong University\\
  {\tt \{linyunfeng,alex-li,wanghanjingwhj\}@sjtu.edu.cn}\\
}

\maketitle

\section*{\centering Abstract}
\textit{
  Visualizing the perceptual content by analyzing human functional magnetic resonance imaging (fMRI) has been an active research area. However, due to its high dimensionality, complex dimensional structure, and small number of samples available, reconstructing realistic images from fMRI remains challenging. Recently with the development of convolutional neural network (CNN) and generative adversarial network (GAN), mapping multi-voxel fMRI data to complex, realistic images has been made possible. In this paper, we propose a model, \textbf{DCNN-GAN}, by combining a reconstruction network and GAN. We utilize the CNN for hierarchical feature extraction and the DCNN-GAN to reconstruct more realistic images. Extensive experiments have been conducted, showing that our method outperforms previous works, regarding reconstruction quality and computational cost.
}

\section{Introduction}

The externalization of the mental content is a fundamental research area in neuroscience. In the last decade, the analysis of multi-voxel fMRI patterns using machine learning techniques allows for the interpretation of visual content. Recent work has progressed from matching seen images to exemplars\cite{NISHIMOTO20111641}, to introducing deep neural networks (DNN)\cite{2015arXiv151006479H} to extract the hierarchical neural representations of the human visual system and to reconstruct the images seen by the subject. 

However, while the brain activity measured by fMRI can be decoded (translated) into DNN features across multiple layers of the network\cite{Shen240317}, the large size of the features, and the absence of regularization in the regression model, contribute to the low decoding accuracy. As a result, the reconstructed image bears little resemblance to the original one. Some model only optimizes the reconstructed image to be similar to natural ones, without utilizing the categorical information of the image. Other work focuses on the reconstruction of a particular type of images\cite{2018arXiv181003856V}\cite{COWEN201412}, which has improved detail but lacked generality.

In this paper, we propose DCNN-GAN, a new model that reconstructs more realistic images from fMRI. Our model consists of a reconstruction network and a recently proposed GAN, pix2pix \cite{isola2017image}, that allows for pixel-wise image generation. In our proposal, an encoder network based on VGG-19\cite{DBLP:journals/corr/SimonyanZ14a} extracts the features from the input image. An fMRI decoder learns the mapping from the fMRI data to the extracted features and decodes features from fMRI test data. In the DCNN-GAN, the reconstruction network outputs the coarse image from the decoded features. The GAN generates a more realistic image from the coarse one.

Our proposed method significantly reduces the size of the decoded features and uses Ridge regression in the decoder to improve numerical stability which contributes to the improvement in decoding accuracy. The DCNN-GAN can render images with more semantically plausible details due to the introduction of category-specific prior. Compared to the reconstruction of a particular image category, our work can be applied to reconstructing various categories of images. Moreover, our proposal stands out for its efficiency, in that it achieves real-time reconstruction.

Through both quantitative comparison and human assessment of the images reconstructed using our method, we have observed an enhancement of the reconstructed image quality among various image categories.
    
\section{Related Work}

As a technically challenging task, reconstructing realistic images from fMRI has been an active area of research in computational neuroscience over the last decades. Before the introduction of deep neural networks(DNN), previous works have only achieved matching the images to similar ones\cite{NISHIMOTO20111641} and reconstruction of contrast-based image\cite{MIYAWAKI2008915} that is low in resolution. These methods directly decode the fMRI into the image to be reconstructed, which limit the number of possible outputs and is unfit for reconstructing images with higher resolution. Some model\cite{Han214247} has used variational autoencoder instead of DNN, resulting in generating relatively blurry images.

Recent works have used DNN\cite{2015arXiv151006479H}\cite{Shen240317}\cite{doi:10.1093/cercor/bhx268} to obtain the features of the input images. With DNN, the process of reconstruction usually involves two crucial steps, the decoding of fMRI and the reconstruction of the image using the decoded DNN features.

{\bf Decoding of fMRI.} The decoding of fMRI activity aims to translate the fMRI pattern measured when the human test subject sees an image, into DNN features that can represent the seen image. Recent work has used sparse linear regression (SLR)\cite{doi:10.1162/089976601750265045} to learn the relationship between fMRI data and the DNN features of all convolutional layers of a VGG-19 model with the same input image\cite{2015arXiv151006479H}. However, we find such linear regression model to behave inadequately without regularization. Also, it is difficult to accurately decode the features of all convolutional layers because the accumulative feature size is too large compared to the small number of fMRI samples.

{\bf Reconstruction of images.} Various methods and models have been introduced to reconstruct images from decoded DNN features. Recent work\cite{Shen240317} has proposed an iterative algorithm that optimizes the reconstructed image so that the DNN features of the image are similar to those decoded from fMRI activity. While the decoded features capture the hierarchical visual information of the image, the difficulty in decoding all features across multiple layers of the VGG-19 model prohibits the model from generating higher quality images. The reconstructed image contains shape that resembles the original image but presents no identifiable textures. We also find the iterative method converges slowly and is therefore unfit for real-time reconstruction.

\section{Approach}
    
\subsection{Neuroscience backgrounds}

    {\bf Human visual system.} 
    The human visual system processes and interprets the visual input to build a mental image of the surroundings. The visual cortex (VC) located in the cerebral cortex is responsible for processing the visual image. It is composed of subsequent regions, namely V1, V2, V3, V4\cite{engel1994fmri}\cite{sereno1995borders}, lateral occipital complex (LOC)\cite{kourtzi2000cortical}, fusiform face area (FFA)\cite{kanwisher1997fusiform}, and parahippocampal place area (PPA)\cite{epstein1998cortical}. In this paper, the entire VC is selected as the brain region of interest (ROI) in the decoding process.

    {\bf Functional magnetic resonance imaging.}
    The fMRI data we used in this paper records the blood-oxygen-level-dependent (BOLD) signal, which measures the hemodynamic response that reflects the level of brain activity\cite{1990PNAS...87.9868O}. As depicted in Figure 1., the fMRI data is a four-dimensional sequence consisting of 3-D volumes sampled every few seconds. Each volume consists of over 700,000 voxels, each voxel measuring $2\times2\times2mm$.
    
    \begin{figure}[h]
        \begin{center}
        \includegraphics[width=85mm]{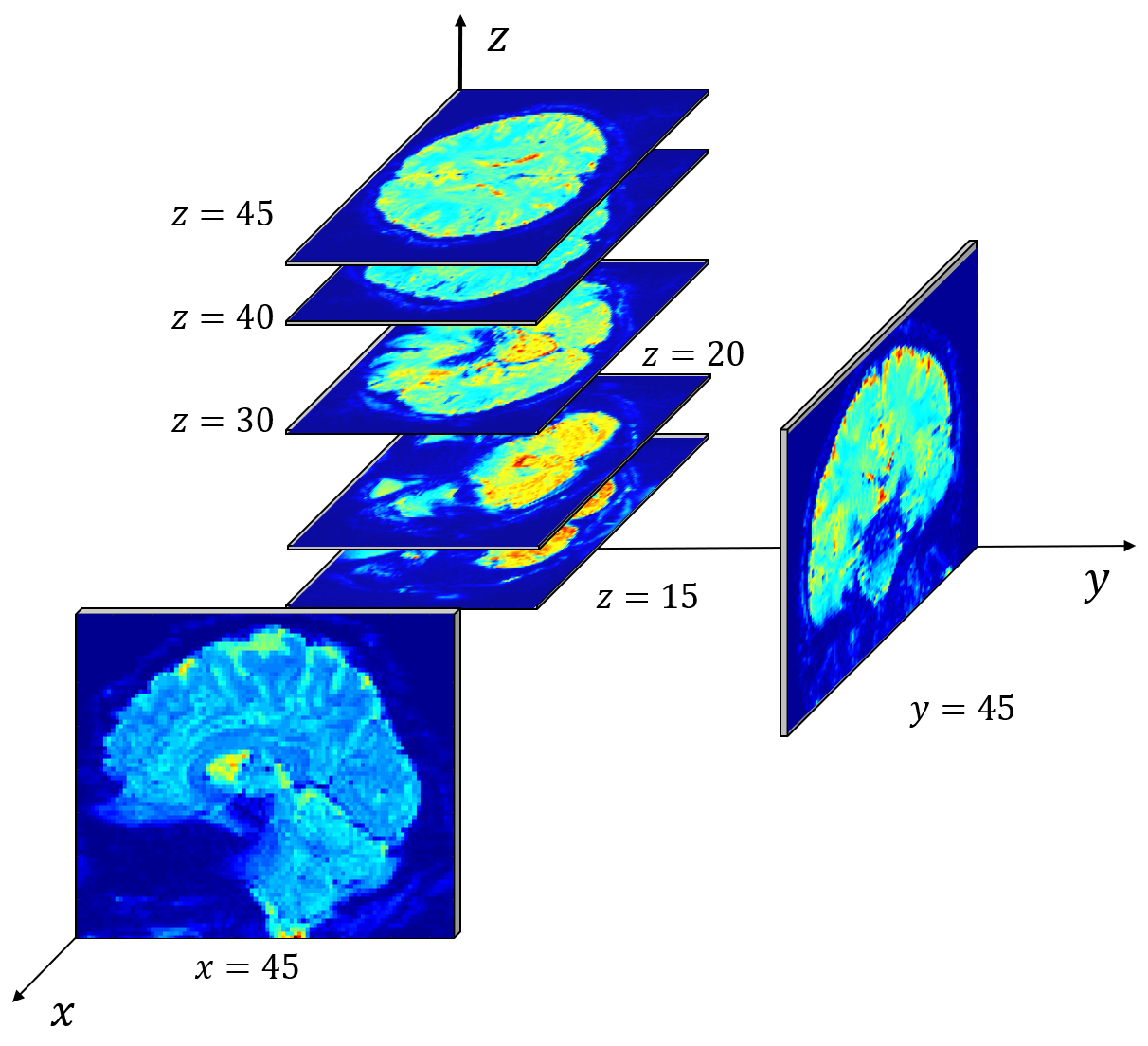}
        \end{center}
        \vspace{1em}
        \caption{Visualization of raw fMRI data, showing slices of a fMRI volume scanned in three directions: sagittal (x), coronal (y) and axial (z).}
        \label{sample-figure}
    \end{figure}
    
    \subsection{Model Formulation}
    
    \begin{figure*}[h]
        \begin{center}
        \includegraphics[width=160mm]{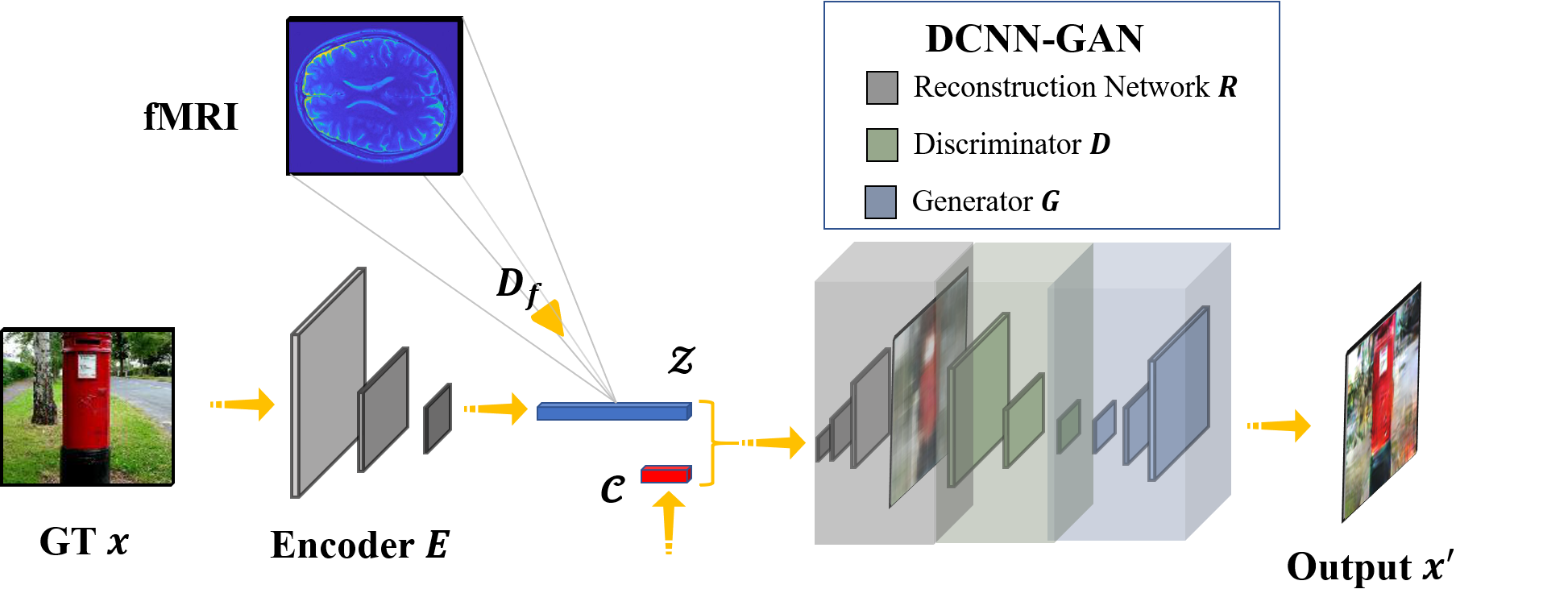}
        \end{center}
        \vspace{1em}
        \caption{Overview of the reconstruction method. The model consists of: 1) The encoder network; 2) The fMRI decoder $D_f$; 3) The DCNN-GAN. Please see Section 3.2 for detailed descriptions.}
        \label{sample-figure}
    \end{figure*}

    In this section, we explain in detail the proposed reconstruction model. As shown in Figure 2, the reconstruction model contains three parts: 
    \begin{enumerate}
        \item The encoder network E, which extracts feature $z$ from the original image $x$.
        \item The fMRI decoder $D_f$, which is trained to learn the mapping from fMRI data to $z$.
        \item The DCNN-GAN that performs image reconstruction using the decoded features.
    \end{enumerate}
    
    
    
    {\bf The encoder network.}
    We build the encoding network E using VGG-19 model pre-trained on ILSVRC2012\cite{ILSVRC15}. By exploiting the feature extracting property of the VGG-19, E maps the original image $x$ to feature vector $z$, as well as the categorical information $c$ of the image. In order to reduce the computational cost of training the fMRI decoder and to improve the accuracy of decoding, we take the output of the first fully connected layer (fc\_7) as the feature vector $z$. This will reduce the dimension of $z$ to $4096$, which is sufficient for preserving the visual information of the original image.
    
    
    {\bf The fMRI decoder.}
     We use a linear least-squares regression model with Tikhonov regularization (Ridge regression)\cite{Hoerl:2000:RRB:338441.338461} to build the decoder. In previous work, an ordinary least squares model is used, which is unstable due to the absence of regularization. The regularization technique increases the numerical stability of our model.
    
    Given the feature $z$ of an input image, and the fMRI multi-voxel data $X$ when showing subject the same image, the model computes the weight vector $w$ and the bias $b$ in the regression function.
    
    \begin{equation}
        z=w^TX+b
    \end{equation}
    
    By minimizing the objective function
    
    \begin{equation}
        \|z - (Xw + b)\|^2_2 + \alpha \|w\|^2_2
    \end{equation}
    
    where $\|w\|^2_2$ is the L2 regularization term, and $\alpha$ is the regularization strength parameter.
    
    {\bf DCNN-GAN.}
    DCNN-GAN is defined as a combination of the reconstruction network R and GAN. R is a deconvolution network that reconstructs the coarse image from decoded feature vector $z$. The GAN, composed of a generative network G and a discriminative network D, takes a coarse image $R(z)$ and the categorical information $c$ as input and outputs a refined image $x'$.
    
    
    The idea behind is that the information decoded from fMRI is insufficient to reconstruct realistic images, and rendering semantically essential details to the image without knowing its category is impractical. Therefore, we use GAN to introduce the image prior, based on the categorical information of the image. This will optimize the reconstructed image to be similar to images of the same category, adding more semantically plausible details to the image.

    The objective function of the network is the combination of the reconstruction network loss, the conditional GAN loss, and a traditional L1 loss. The generative network tries to minimize the objective while the discriminative network tries to maximize it. Therefore, the final objective function can be expressed as:
    
    
    \begin{equation}
        \begin{aligned}
        G=\arg \min_C \max_D \mathcal{L}_{cGAN}(G,D) + \lambda\mathcal{L}_{L1}(G)\\ 
        +\theta\mathcal{L}_{2}(R)
        \end{aligned}
    \end{equation}
    
    Given $w$ the random noise, 
    
    \begin{equation}
        \begin{aligned}
        \mathcal{L}_{cGAN}(G,D)=\log D(R(z), x') \\+
        \log (1-D(R(z),G(R(z),w)))
        \end{aligned}
    \end{equation}
    
    is the loss function of the conditional GAN, and
    
    \begin{equation}
        \mathcal{L}_{L1}(C)=\|x'-G(R(z),w)\|_1
    \end{equation}
    
    is the L1 loss function, and
    
    \begin{equation}
        \mathcal{L}_{R}(G)=\|x-R(z)\|_2
    \end{equation}
    
    is the loss function of the reconstruction network. We optimize the loss of reconstruction network and the loss of $G,D$ alternatively in practice.

    
\section{Experiment}

In this section, we validate the effectiveness of our reconstruction model by experiments. We have trained and tested our model on the following datasets. 

\subsection{Datasets} 

\textbf{ILSVRC2012.} The ILSVRC2012 dataset is a subset of the large hand-labeled ImageNet dataset. The validation and test data consist of 150,000 images, and the training data contains 1.2 million images. All the images are hand labeled with the presence or absence of 1000 object categories which do not overlap with each other. 

\textbf{fMRI on ImageNet.} Originally used in \cite{2015arXiv151006479H}. The fMRI data were recorded while subjects were viewing object images (image presentation experiment) or were imagining object images (imagery experiment). In the training image session, a total of 1,200 images from 150 object categories were each presented only once. In the test image session, a total of 50 images from 50 object categories were presented 35 times each. All images were from ImageNet (Fall 2011 release). In this paper, we use the fMRI data of the VC region in the image presentation experiment. The number of voxels in the VC region is 4466. The number of fMRI samples available is 2700.

\subsection{Implementation Details}
We have implemented our approach using Pytorch \cite{paszke2017automatic}.
The encoder network E is based on VGG-19 model, pretrained on ILSVRC2012 training set. The input of E is the original image of size $224 \times 224 \times 3$. We take the output of the first fully connected layer of VGG-19 with size $4096 \times 1$ as the feature vector used to train the fMRI decoder.

The fMRI decoder is constructed using a linear regression model. We have chosen Ridge regression, which is a linear least squares with L2 normalization. The number of samples in the training set is $~2200$, and the number of fMRI voxels is $~4400$. The L2 regularization strength $\alpha$ is empirically set to $0.7$.

The reconstruction network R is a deconvolution network consisting of a fully connected layer with input size $4096 \times 1$ and output size $512 \times 7 \times 7$, followed by 4 deconvolutional layers with kernel size set to $4\times4$ and output channel set to $256, 128, 128, 128$, and a convolutional layer with kernel size set to $1\times1$ and output channels set to $3$. The output size of G is $112\times112\times3$. We used the Adam optimizer, with initial learning rate set to 0.01 and exponential decay. On the ILSVRC2012 training set, we trained our network for 200 epochs with batch size set to 256. We add Gaussian noise to the input to increase the robustness of the network.

We trained the GAN using the output images from the reconstruction network R and the corresponding original images as paired data on a specific category of the ILSVRC2012 training set for 500 epochs. The input is resized to $128\times128\times3$, the same size as the output, the batch size is 256, and the learning rate is 0.001.


\subsection{Visual comparison with previous models}

\begin{figure}[h]
  \includegraphics[width=85mm]{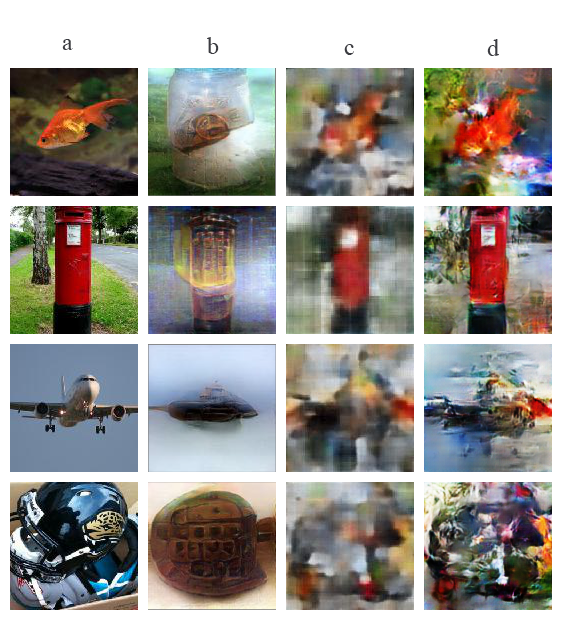}
  \vspace{1em}
  \caption{Comparison of reconstructed images from different models. (a) Original images (b) Reconstructed images (previous model) (c) Reconstructed images (the output of R) (d) Reconstructed images (the output of DCNN-GAN)}
  \label{sample-figure}
\end{figure}

The results of the existing model are less recognizable because they tend to preserve the shape of the object while sacrificing the texture, while our proposal reaches a balance between the shape and texture. Some of the test instances (i.e. the mailbox) introduce objects that do not belong to the input, indicating that the existing model may over-fit the training set. From the comparison we can conclude that, the outputs of the DCNN-GAN are significantly improved in resolution and details when compared to the intermediate outputs of the deconvolution network. It supports the idea that GAN can optimize the 
generation of images by introducing category-specified prior knowledge.


\subsection{Comparison of decoding accuracy}

In this part, we compare the fMRI decoding accuracy between the linear regression model used in previous work and the ridge regression model in our method. We use two common regression metrics. The coefficient of determination, as known as R-squared, reflects the goodness of fit of a model, where 1 is the best and lower the worse. The root-mean-square error (RMSE) measures the error between the predicted and observed values. The result shown in Table 1 indicates an improvement in decoding accuracy using our model compared to the previous model.

\begin{table}[h]
  \caption{Results of decoding accuracy}
  \begin{center}
    \begin{tabular}{c | c c c}
      \hline
      \hline
      \makebox[10mm]{Metrics} & \makebox[10mm]{Linear} & 
      \makebox[10mm]{Ridge} & \makebox[10mm]{Mean}\\
      \hline
      R-squared & -0.3093 & 0.3184 & -0.0039 \\
      RMSE & 0.4960 & 0.3614 & 0.4402 \\
      \hline
      \hline
    \end{tabular}
    \label{sample-table}
  \end{center}
\end{table}

\subsection{Human Perceptual Study}


Evaluating the quality of the reconstructed images is an open problem. None of the traditional metrics can effectively measure the similarity between two images with complex structural and textural information. Therefore we conducted a double-blind perceptual study on a group of randomly selected volunteers in the interest of holistic evaluation of our results. 

{\bf Perceptual study.} Over $40$ volunteers were surveyed in the study, and each of them was presented a sequence of trials on which the image reconstructed by the existing model was pitted against the image generated by our model with the same input. Given the origin image $x$, volunteers were asked to select the image which they viewed as a better reconstruction of $x$ and report whether it was a satisfying reconstruction. Each image was presented for $1$ second, and volunteers were given unlimited time to decide their response.

The result of the perceptual is listed as follows.
\begin{table}[h]
  \caption{Results of perceptual study}
  \begin{center}
    \begin{tabular}{c | c c}
      \hline
      \hline
      \makebox[13mm]{Items} & \makebox[20mm]{Existing model} & 
      \makebox[20mm]{Our proposal}\\
      \hline
      Reconstruct(\%) & 44.3 & 55.7\\
      Satisfaction(\%) & 12.4 & 20.1\\
      \hline
      \hline
    \end{tabular}
    \label{sample-table}
  \end{center}
\end{table}

We can derived the conclusion that our model has achieved improvement in performance compared to existing work.

\section{Conclusion}
In this paper, we proposed a novel model DCNN-GAN and applied it to the output features of the fMRI decoder to reconstruct realistic images. Compared to previous works, the overall quality of the reconstructed image was considerably enhanced, and our model achieved real-time reconstruction. The future work is to reconstruct realistic images of which categories do not exist in the training set.
\renewcommand\refname{Reference}
\bibliographystyle{unsrt}
\bibliography{cite}

\end{document}